\documentclass[11pt,a4paper]{article}
\usepackage[hyperref]{conll2018}
\usepackage{times}
\usepackage{latexsym}
\usepackage{subcaption}
\usepackage{graphicx}
\usepackage{footnote}
\makesavenoteenv{tabular}
\makesavenoteenv{table}

\usepackage{subcaption}
\usepackage{url}

\aclfinalcopy 

\title{From Random to Supervised: A Novel Dropout Mechanism Integrated with Global Information}

\makeatletter
\def\thanks#1{\protected@xdef\@thanks{\@thanks
        \protect\footnotetext{#1}}}
\makeatother

\author{Hengru Xu\textsuperscript{1, \dag\thanks{$^{\dag}$ Hengru Xu and Shen Li contributed equally to this work.}{}} \And
  Shen Li\textsuperscript{2, \dag} \And
  Renfen Hu\textsuperscript{3} \And
  Si Li\textsuperscript{1} \And
  Sheng Gao\textsuperscript{1, \ddag\thanks{$^{\ddag}$ Corresponding author.}{}} \AND
  {\tt \textsuperscript{1}\{xuhengru, lisi, gaosheng\}@bupt.edu.cn} \\
  {\tt \textsuperscript{2}shen@deeplycurious.ai} \\
  {\tt \textsuperscript{3}irishere@mail.bnu.edu.cn} \\
  \textsuperscript{1} SICE, Beijing University of Post and Telecommunication \\
  \textsuperscript{2} Deeplycurious.ai \\
  \textsuperscript{3} Institute of Chinese Information Processing, Beijing Normal University \\
}

\date{}

\begin{document}
\maketitle
\begin{abstract}
Dropout is used to avoid overfitting by randomly dropping units from the neural networks during training. 
Inspired by dropout, this paper presents GI-Dropout, a novel dropout method integrating with global information to improve neural networks for text classification. 
Unlike the traditional dropout method in which the units are dropped randomly according to the same probability, we aim to use explicit instructions based on global information of the dataset to guide the training process.
With GI-Dropout, the model is supposed to pay more attention to inapparent features or patterns. 
Experiments demonstrate the effectiveness of the dropout with global information on seven text classification tasks, including sentiment analysis and topic classification.
\end{abstract}

\section{Introduction}
Recently, neural networks have achieved remarkable results in natural language processing (NLP).
Convolutional Neural Network (CNN) and Recurrent Neural Network (RNN) are two popular types of neural network architectures and both of them are widely applied to various NLP tasks.
CNN is known for its strong ability in extracting position-invariant features and RNN is highlighted in modeling sequences \cite{yin2017comparative}.
In sentence classification tasks, models based on CNN or RNN aim to represent sentences as appropriate embeddings, which are supposed to encode semantic features for the classification.

However, with the consideration of computational complexity and spatial limitation, neural networks are often trained via mini-batch in which global information is gathered implicitly rather than explicitly. To facilitate the learning process, \newcite{li2017initializing} extract global semantic features from the training dataset, and encode them into CNN filters with a novel initialization mechanism. This approach gains significant improvements in sentiment analysis and topic classification tasks.

Unlike most of machine learning methods, the advantage of neural networks is extracting features with less need of feature engineering. In general, the stronger ability of a model to learn features automatically, the better performance it will achieve. However, during the training process, neural networks tend to focus on some distinctive words or phrases but ignore other noteworthy patterns, which may result in overfitting, especially in a small dataset. To avoid this problem, dropout is proposed \cite{hinton2012improving,srivastava2014dropout}. The key idea of dropout is to randomly drop units from the neural network during training and use a smaller weight of these units in the test. 

Inspired by the above works, we propose a novel dropout method guided by global information (GI-Dropout). In our method, we force the model to pay more attention to features that are inapparent or with low frequency by dropping words that are prominent and easy to learn. Unlike the traditional dropout method where neurons are dropped randomly with the same probability, we encode global information into dropout. Specifically, we drop words based on their importance which are calculated from training data via a novel Na\"ive Bayes (NB) weighting technique.

With this dropout method, neural networks tend to extract not only the obvious features but also the unobvious features which are also helpful for the classification. 
By integrating our method into a classic CNN model for text classification \cite{kim2014convolutional} and a novel self-attentive RNN \cite{lin2017structured}, we observe significant improvements in various benchmarks.\footnote{We release source codes at \href{https://gitlab.com/xusong19960424/global_cnn}{https://gitlab.com/xusong19\\960424/global\_cnn}.}
The advantages of our approach are as follows:
\begin{enumerate}
\item Global information is directly obtained from the training data without any external resources; 
\item GI-Dropout is simple but effective, and could be easily applied to other DNN models;
\item The computation brought by our method is relatively small, resulting in little additional training cost.
\end{enumerate}

\section{Related Work}
Recently, neural networks dominate the state-of-the-art results on a wide range of NLP tasks. For text classification, 
\newcite{kim2014convolutional} proposes a classical one-layer CNN which is very efficient for feature extraction, and it is considered as a strong baseline for various sentiment and topic classification tasks. Following this work, \newcite{yin2015multichannel} introduce multichannel variable-size convolution, and \newcite{zhang2016mgnc} exploit different pre-trained word embeddings (e.g. word2vec and GloVe). \newcite{zhang2017sensitivity} improve the CNN model by optimizing hyper-parameters and provide a detailed sensitivity analysis. 

RNNs also achieve comparable performance in this area. \newcite{tang2015document} show that gated RNN performs well on document-level sentiment classification.
\newcite{lin2017structured} propose a enhanced model to extract an interpretable sentence embedding by introducing self-attention mechanism and yields a significant performance gain compared with other sentence embedding methods.

\newcite{yin2017comparative} make a systematic comparison of CNNs and RNNs, showing that both of the networks can provide complementary information for text classification tasks, while which architecture performs better depends on how important it is to semantically understand the global/long-range semantics.

To improve the semantic understanding abilities of the models, some works aim to encode prior knowledge into the networks. For example, \newcite{hu2016harnessing} present a framework that encapsulates the logical structured knowledge into a neural network. 
\newcite{li2017initializing} encode global semantic features into the convolutional filters instead of initializing them randomly, which helps the filters focus on learning useful n-grams.

Another effective method to facilitate learning process is to exploit dropout mechanism. Apparently, if a model pays too much attention to a few distinct patterns, it can easily give rise to an overfitting, especially in a small dataset. \newcite{hinton2012improving} introduce Binary (regular) Dropout, showing that it can prevent co-adaptation of neurons by randomly dropping units from the neural networks during training, so as to reduce overfitting.
Later \newcite{srivastava2014dropout} show that multiplying outputs of the neurons by a random variable drawn from Gaussian distributions works as well, or perhaps better than regular dropout.
\newcite{NIPS2013_5032} present standout, an adaptive dropout method, where each variable's dropout probability is calculated by a binary belief network, which can be trained jointly with the neural networks. 
\newcite{NIPS2015_5666} introduce variational dropout, a generalization of Gaussian dropout where the dropout rates are also learned during training.

The existing dropout methods are often based on mathematics or learned jointly with the downstream task, where global information is not explicitly utilized.
Different from previous works, we focus on how to utilize global information to help model training via dropout.
As depicted in Figure \ref{fig:model}, GI-Dropout is introduced at the beginning of the baseline models, which is different from prior dropout methods which aim at controlling units in the networks rather than input words in the texts.

In this work, we use the global information to guide dropout method by dropping words based on their importance. Hence, neural networks are able to extract not only the obvious features but also the unobvious features which are also helpful for the classification.

\section{Our method}

The intuition behind our method is straightforward. 
Since neural networks aim to capture semantic features and classify sentences by the features, we encourage models to share more attention to unobvious features by dropping words according to their importance.
Some features are so distinctive that model can learn them easily. However, a sentence may have more than one feature that can contribute to class prediction. 
For instance, in ``The story is sad and very boring'', ``boring'' is of strong polarity and indicates negative emotion.
Neural networks may not be sensitive to other features like ``sad'' which is also helpful for the sentiment classification, due to the very strong impact of ``boring''.
In GI-Dropout, a word of higher importance score has greater possibility to be dropped.
Thus, models are forced to learn unobvious features and will achieve better performance in prediction.

\subsection{Importance Score}
\label{Importance Score}
Firstly, we compute an importance score for each word.
Intuitively, word ``unique'' is much more important than ``movie'' for determining polarities of reviews. 
Na\"ive Bayes (NB) weighting is an effective technique for determining the importance of words \cite{martineau2009delta, wang2012baselines, li2017initializing}.
The NB weight $\emph{r}$ of word $w$ in class $c$ is calculated as follows:

\begin{equation}\label{tra_nb}
 \emph{r}_{c}^{w} = \frac{(\emph{n}_{c}^{w} + \alpha ) /\left \| \emph{n}_{c} \right \|_{1}} {(\emph{n}_{\tilde{c}}^{w} + \alpha )/\left \| \emph{n}_{\tilde{c}} \right \|_{1}}
\end{equation}
where $\emph{n}_{c}^{w}$ is the count of word $w$ in class $c$,
$\emph{n}_{\tilde{c}}^{w}$ is the count of word $w$ in the other classes,
$\left \| \emph{n}_{c} \right \|_{1}$ is the count of all the word occurrences in class $c$,
$\left \| \emph{n}_{\tilde{c}} \right \|_{1}$ is the count of all the word occurrences in the other classes,
$\alpha$ is a smoothing parameter and is set as 1 in this paper. 

To avoid low-frequency words being recognized as important words, we propose an improved NB weighting method based on (\ref{tra_nb}):

\begin{equation}\label{new_nb}
  \emph{r}_{c}^{w} = \frac{(\emph{n}_{c}^{w} + \alpha ) /\left \| \emph{n}_{c} \right \|_{1}} {(\emph{n}_{\tilde{c}}^{w} + \alpha )/\left \| \emph{n}_{\tilde{c}} \right \|_{1}} \times \log_{\beta}\emph{n}_{c}^{w}
\end{equation}

where $\log_{\beta}\emph{n}_{c}^{w}$ is introduced as a frequency factor. The base $\beta$ is a hyperparameter.

\begin{figure}[!t]
\centering
\includegraphics[width=\linewidth]{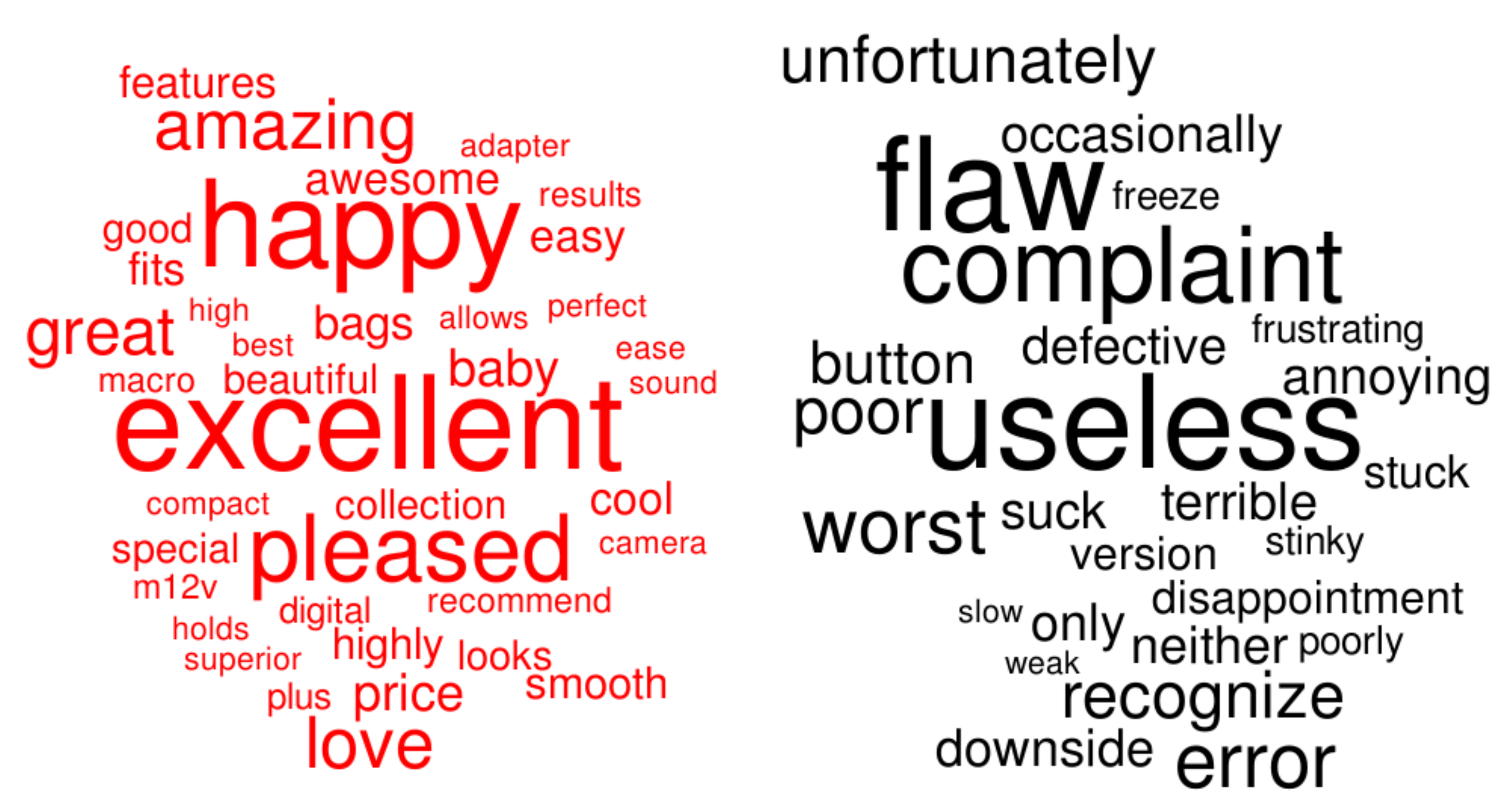}
\caption{Top 30 key words of each class in Customer Review dataset}
\label{fig:keyword}
\end{figure}

For positive class in movie review dataset (MR), the scores of words like ``unique'' and ``warm''
should be large since they appear much more frequently in positive texts than in negative texts.
As for neutral words like ``the'' and ``movie'', their scores should be small. 
For a word $w$, we select the max score of it as its importance score:
\begin{equation}\label{final_score}
  \emph{r}^{w} = max(\emph{r}_{c_{0}}^{w}, \emph{r}_{c_{1}}^{w}, ..., \emph{r}_{c_{n}}^{w})
\end{equation}

In Figure \ref{fig:keyword}, we show top 30 key words of each class in customer review dataset (CR). 
We aim to drop these key words with higher probabilities and encourage the model to pay more attention to other unobvious features. 

\subsection{Dropout Probability}

\begin{figure}[!t]
\centering
  \includegraphics[width=\linewidth]{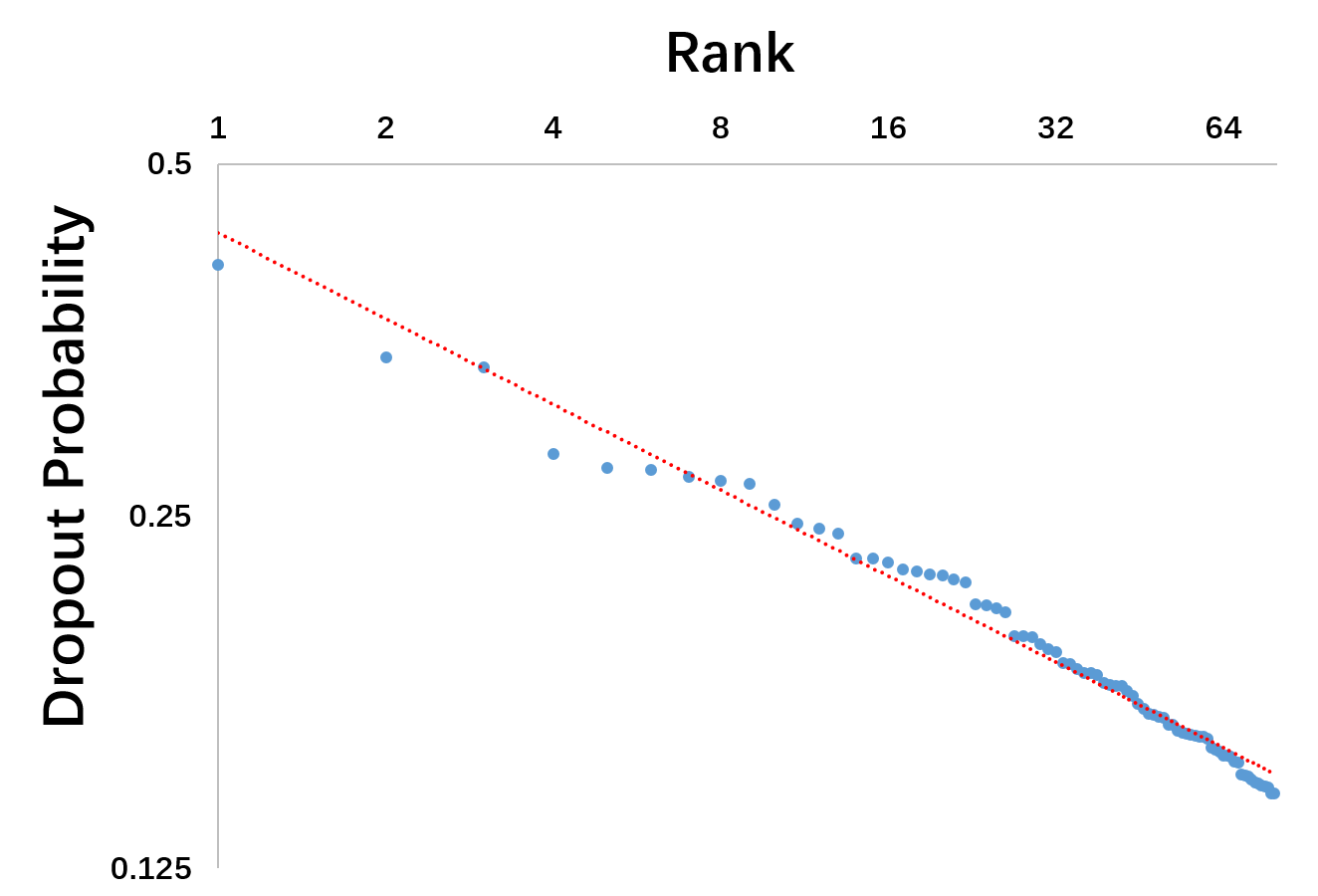}
  \caption{GI-Dropout probability and rank in SST-1 with $\beta = 0.95$.}
  \label{fig:zipf}
\end{figure}

As shown in \ref{Importance Score}, we compute words importance scores with the whole training data. It is a simple yet effective way to represent the global information. After obtaining the scores, we compress them into $[0, 1)$. The GI-Dropout probability of word $w$ is:
\begin{equation}
\label{equation:our fuction}
p(r) = \frac{e^{r} - 1}{e^{r} + 1}
\end{equation}
where $r$ is the importance score of $w$ calculated via (\ref{new_nb}). A word would not be ignored when its probability is 0.

The $\beta$ in (\ref{new_nb}) is a key parameter. 
As shown in Figure \ref{fig:zipf}, after tuning $\beta$, the GI-Dropout probability of a word and its probability rank follow Zipf's Law. 
\newcite{zipf1935psychology} states that given a sample of words, the frequency of any word is inversely proportional to its rank in the frequency table.
Replacing the frequency with GI-Dropout probability, we can get a variant of Zipf's Law.
The experiments will show that setting $\beta$ to this value in SST-1 is not a coincidence.

\begin{figure}[!t]
\centering
  \includegraphics[width=\linewidth]{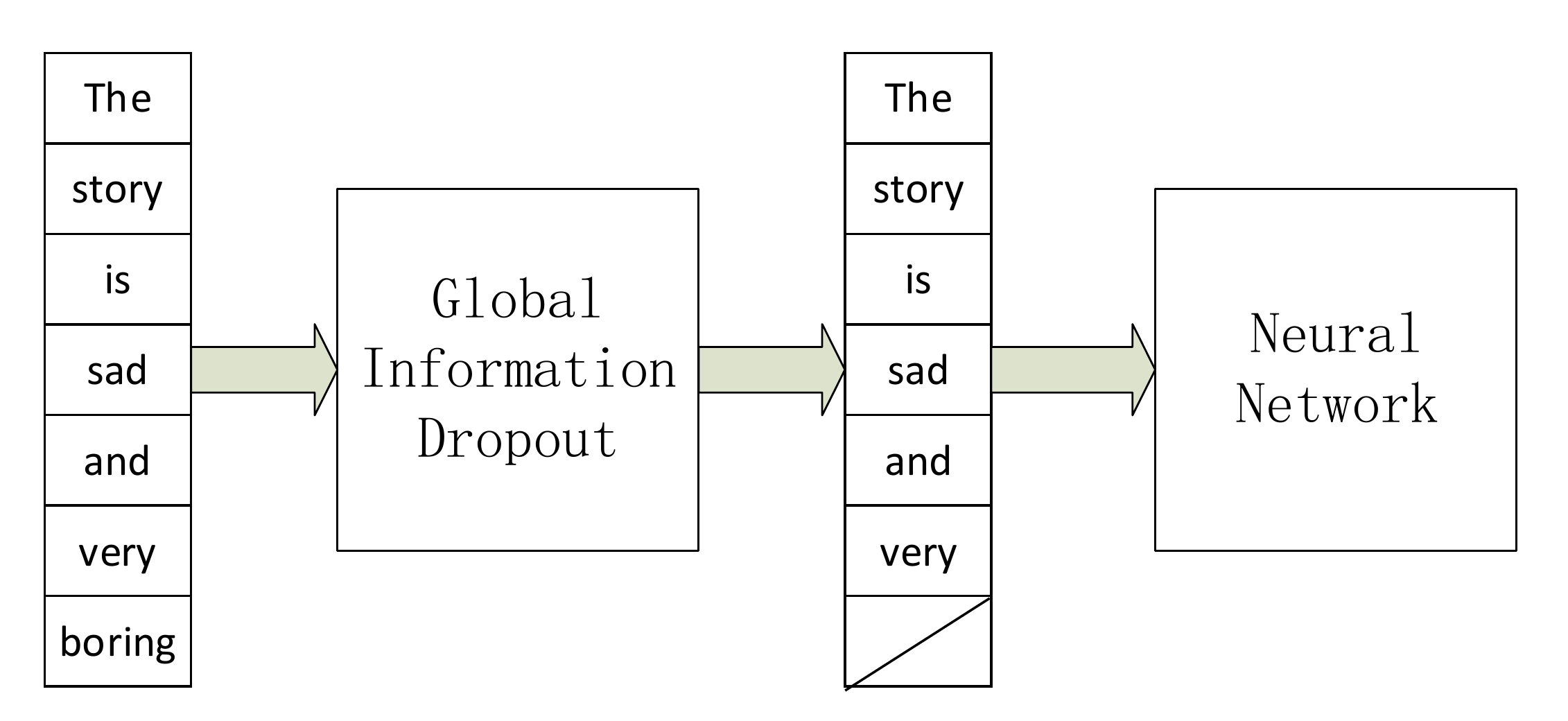}
  \caption{GI-Dropout. In this case, the word embedding of ``boring'' is dropped and set to zero vector while ``sad'' is not.}
  \label{fig:model}
\end{figure}

\subsection{GI-Dropout Method}
As illustrated in Figure \ref{fig:model}, we implement a GI-Dropout layer before the neural network.
Models without our dropout method can be viewed as the special case in which all the words are not dropped in GI-Dropout layer, i.e. dropout probabilities of all words are 0. 

In this paper, every word in training data has a score to measure its importance via the novel NB weighting method, as well as a dropout probability calculated by the proposed scale function.
During training, the words will be dropped according to their dropout probabilities.

The way to implement our dropout method is very straightforward. 
In embedding layer, we get the word embedding $e_{i}$ of word $w_{i}$ after looking it up in the embedding table. After that, this word can be dropped according to its GI-Dropout possibility. For word $w_{i}$, we set the $e_{i}$ to zero vector if it needs to be dropped.
Through this method, the neural network will not learn features from words whose embeddings are zero vectors. 
It is worth noted that the dropout probabilities of words differ from each other, which is different from the traditional dropout method where all the neurons are dropped according to the same probability.
The dropout probabilities which are encoded with global information, guide the model to share attention to unobvious patterns. 
 
In traditional dropout method, all the neurons are used in testing, but their weights are scaled down by a factor $p$ (same with $p$ in training) since a part of units emit nothing to the next layer during training.
While in our method, during evaluation and testing, dropout probabilities of all the words are set to 0 so as to use all the patterns, and scaling is not needed.

\section{Experiments}
CNN-non-static proposed by \newcite{kim2014convolutional} is considered as a very strong baseline in sentence classification.
Self-attentive RNN proposed by \newcite{lin2017structured} also achieves outstanding performance in many sentence classification tasks. 
We adopt these two models to evaluate GI-dropout.

\subsection{Datasets}

\begin{table}[!t]
\renewcommand{\arraystretch}{1.3}
\centering
\begin{tabular}{c c c c c c}
\hline
\bfseries Dataset & c & l & N & V  & Test\\
\hline
MR & 2 & 20 & 10662 & 18765 &  CV\\
SST-1 & 5 & 18 & 11855 & 17836  & 2210\\
SST-2 & 2 & 19 & 9613 & 16185  & 1821\\
Subj & 2 & 23 & 10000 & 21323  & CV\\
TREC & 6 & 10 & 5952 & 9592  & 500\\
CR & 2 & 19 & 3775 & 5340  & CV\\
MPQA & 2 & 3 & 10606 & 6246 & CV\\
\hline
\end{tabular}
\caption{\ Datasets summary. 
c: Number of target classes. 
l: Average sentence length.
N: Dataset size.
V: Vocabulary size. 
Test: Test set size (CV means there is no standard train/test split and thus 10-fold CV is used).
}
\label{table: dataset summary}

\end{table}

Following \citep{kim2014convolutional}, we evaluate the performance of the proposed approach on various datasets. 
We use the same seven datasets with \citep{kim2014convolutional}, including both sentiment analysis and topic classification tasks:

\textbf{MR}: Movie reviews sentiment datasets\footnote{https://www.cs.cornell.edu/people/pabo/movie-review-data/}.

\textbf{SST-1}: Stanford Sentiment Treebank with 5 sentiment labels \citep{socher2013recursive}\footnote{http://nlp.stanford.edu/sentiment/}. The data consists of phrases-level and sentence-level instances. To keep same with \citep{kim2014convolutional},  we train the model on both phrases and sentences but only test on sentences.

\textbf{SST-2}: SST-1 data with binary labels.

\textbf{Subj}: Subjective or objective classification dataset \citep{pang2005seeing}.

\textbf{TREC}: 6-class question classification dataset \citep{li2002learning} \footnote{http://cogcomp.cs.illinois.edu/Data/QA/QC/}.

\textbf{CR}: Customer products review dataset \citep{hu2004mining} \footnote{http://www.cs.uic.edu/∼liub/FBS/sentiment-analysis.html}.

\textbf{MPQA}: Opinion polarity detection dataset \citep{wiebe2005annotating}.

The statistics of the datasets can be seen in Table \ref{table: dataset summary}.

\subsection{CNN Model}

\begin{figure}[!t]
\centering
  \includegraphics[width=\linewidth]{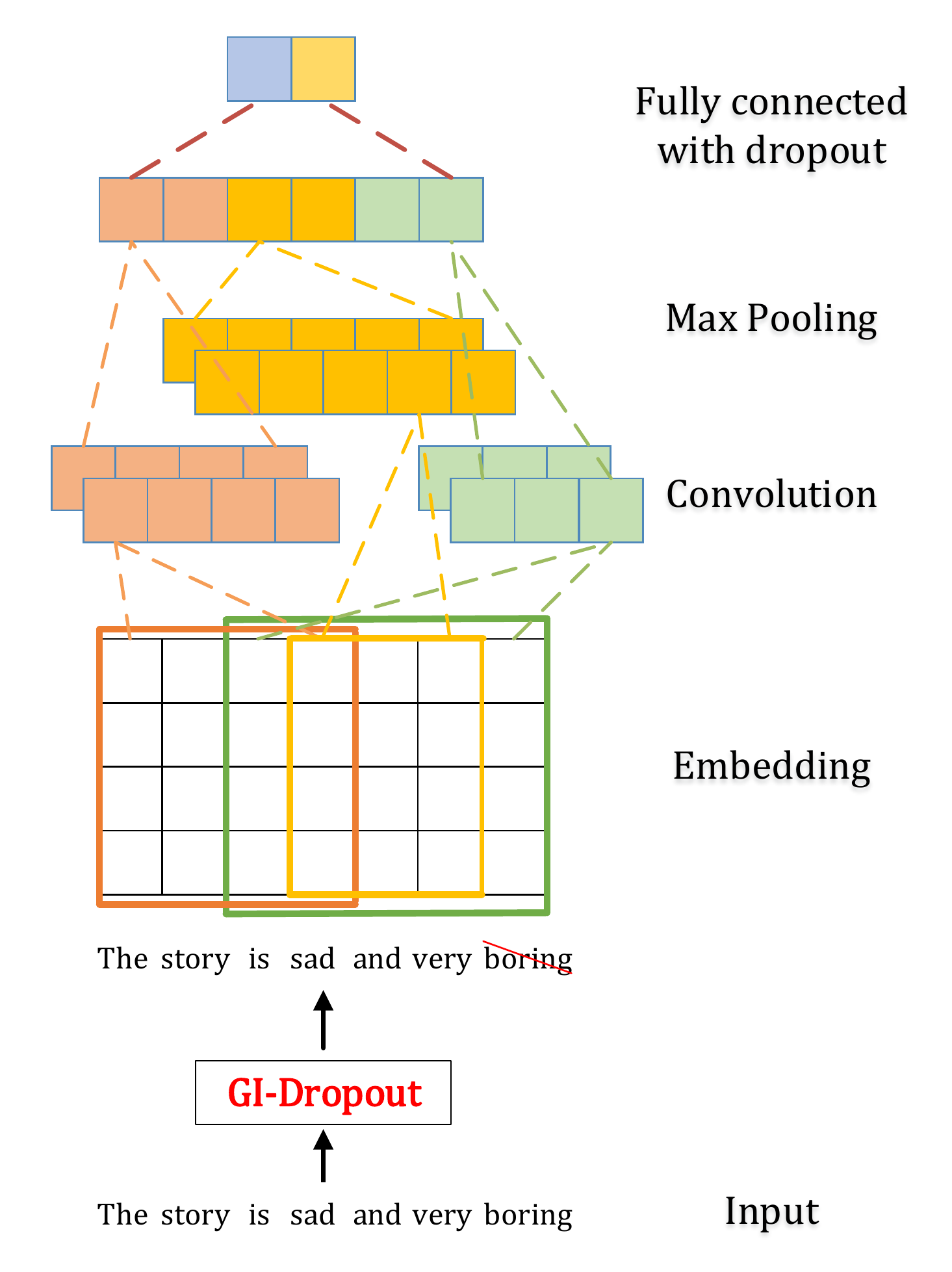}
  \caption{CNN architectures with GI-Dropout.}
  \label{fig:cnn}
\end{figure}

CNNs use filters to capture semantic features of n-grams.
After that, max-pooling is introduced to force the network to capture the most useful local features produced by convolutional layers \cite{collobert2011natural}.
A simple CNN model in \cite{kim2014convolutional} consists of the embedding layer, one convolution and pooling layer, and one fully connected layer. Four model variations are provided in \cite{kim2014convolutional}, and we choose the CNN-non-static model as our baseline. The hyperparameters of the CNN are described in Table \ref{table: baseline hyperparameters}. The architecture of the model integrated with GI-Dropout is shown in Figure \ref{fig:cnn}.

\begin{table}[!t]
\renewcommand{\arraystretch}{1.3}
\centering
\begin{tabular}{c c}
\hline
\bfseries Parameters & \bfseries Values\\
\hline
Word embeddings & GoogleNews-negative300 \footnote{A widely used publicly available word2vec 300-dimension vectors which were trained on 100 billion words from Google News in \cite{mikolov2013distributed} way.} \\
Fine-tune & Yes\\
Convolution & 1-d \\
Filter size & [3, 4, 5] \\
Filter numbers & 300 \\
Activation function & ReLU \\
Pooling method & max-over-time \\
MLP dropout rate & 0.5 \\
\hline
\end{tabular}
\caption{CNN configuration.}
\label{table: baseline hyperparameters}
\end{table}

\subsection{Self-attentive RNN Model}

\begin{figure}[!t]
\centering
  \includegraphics[width=\linewidth]{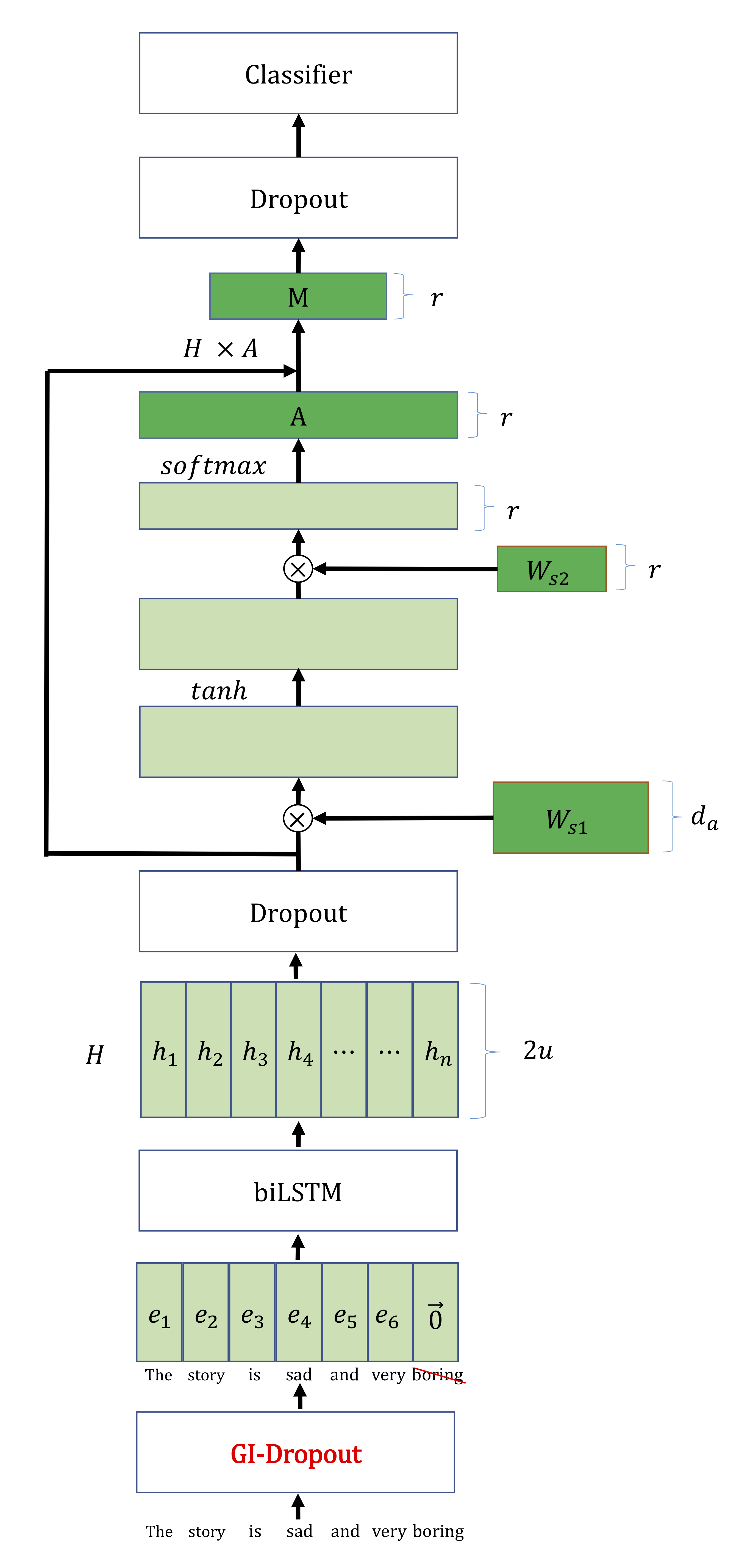}
  \caption{Self-attentive RNN architectures.}
  \label{fig:rnn}
\end{figure}

Long Short-Term Memory (LSTM) is a specific recurrent neural network (RNN) architecture which is good at modeling temporal sequences and can capture long-range dependencies \citep{sak2014long}.
Attention mechanism, first proposed in \citep{bahdanau2014neural}, has become an integral part of sequence modeling.
The self-attentive RNN proposed by \newcite{lin2017structured} consists of a bidirectional LSTM (biLSTM) and the self-attention mechanism.
Self-attention mechanism is used to replace the max pooling or averaging step after the biLSTM.
Multiple hops of attention are performed to extract semantic features in different aspects of the sentence.

In brief, suppose we have a sentence of $n$ tokens, and let the hidden unit number for each unidirectional LSTM be $u$.
After the biLSTM layer, we can get $H$, which have the size of $n$-by-$2u$.
The attention mechanism takes the whole LSTM hidden states $H$ as input, and outputs a vector of weights $a$,
\begin{equation}
\emph{a} = softmax(w_{s2}\tanh(W_{s1}H^T))
\end{equation}
where $W_{s1}$ is a weight matrix with a shape of $d_{a}$-by-$2u$, and $W_{s2}$ is a vector of parameters with size $d_{a}$ which is a hyperparameter.

To extract $r$ different aspects of the sentence, \newcite{lin2017structured} present multiple hops of attention, i.e. extend the $w_{s2}$ into a $r$-by-$d_{a}$ matrix and note it as $W_{s2}$.
In the end, the annotation vector $a$ becomes annotation matrix $A$.
\begin{equation}
\emph{A} = softmax(W_{s2}\tanh(W_{s1}H^T))
\end{equation}

The sentence embedding is:
\begin{equation}
\emph{M} = \emph{A}\emph{H}
\end{equation}

Then the paper uses two layer 2-layer MLP with ReLU activation function to predict the label of the sentence. Besides, a penalization term is introduced to encourage the diversity of summation weight vectors across different hops of attention.

Since \newcite{lin2017structured} do not provide source codes, we reproduce the model and integrate dropout layers into the model as shown in Figure \ref{fig:rnn}. 
We perform a grid search to get the best baseline hyperparameters with which the model can achieve the state-of-the-art accuracy in most of the datasets.
This model uses a bidirectional LSTM with $300$ dimensions of hidden states in each direction.
In self attention part, $d_a$ is $350$ and the coefficient of the penalization term is $1$.
$r$ is set to $4$ considering the size of datasets and the length of texts.
We also use a 2-layer ReLU output MLP with 2000 hidden units.
During training we use a $0.5$ dropout rate on the MLP.
The hyperparameters are described in Table \ref{table: self-attention baseline hyperparameters}.

\begin{table}[!t]
\renewcommand{\arraystretch}{1.3}
\centering
\begin{tabular}{c c}
\hline
\bfseries Parameters & \bfseries Values\\
\hline
Word embeddings & Glove-300 \footnote{A widely used publicly available 300-dimension word embeddings \citep{pennington2014glove}.} \\
Fine-tune & Yes\\
biLSTM hidden units & 300 \\
$d_{a}$ & 350 \\
$r$ & 4 \\
MLP Activation & ReLU \\
MLP dropout rate & 0.5 \\
\hline
\end{tabular}
\caption{Self-attentive RNN configuration.}
\label{table: self-attention baseline hyperparameters}
\end{table}

\subsection{Experiment Settings}
We apply our method to two baseline models.
For fair comparison, we use the same hyperparameters settings with two baselines for training and testing. For datasets that do not have test sets, we split them for cross-validation with fixed random seeds. We train all the models using early stopping and set timedelay to 10.

\newcommand{\specialcell}[2][c]{%
  \begin{tabular}[#1]{@{}c@{}}#2\end{tabular}}
\begin{table*}[t]
\begin{small}
\renewcommand{\arraystretch}{1.3}
\centering
\begin{tabular}{c |ccccccc}
\hline
\bfseries Model             & MR   & SST-1 & SST-2 & Subj & TREC &  CR  & MPQA\\
\hline
CNN-non-static & 81.5 & 48.0  & 87.2  & 93.4 & 93.6 & 84.3 & 89.5\\
\hline
CNN-reproduce       & 81.4 & 47.8  & 87.5  & 93.0 & 92.4 & 84.3 & 89.6\\
CNN-Dropout-same (p)       &  81.5(0.1) & 48.5(0.1)  & 87.6(0.1)  & 93.5(0.2) & 92.9(0.1)& 84.5(0.5) & 87.4(0.1)\\

CNN-GI-Dropout ($\beta$) & \textbf{81.9}(0.87) & \textbf{49.0}(0.95)  & \textbf{88.1}(0.98)  & \textbf{93.4}(0.91) & \textbf{93.2}(0.83) & \textbf{85.1}(0.87) & \textbf{89.8}(0.98)\\

\hline
\hline
RNN-baseline & 82.1 & 49.7  & 89.7  & 93.6 & 92.6 & 84.1 & 89.6\\
RNN-Dropout-same (p) & 82.2(0.2) & 51.9(0.1)  & 90.1(0.1)  & 93.9(0.1) & 93.4(0.2) & 84.2(0.1) & \textbf{89.7}(0.1)\\
RNN-GI-Dropout ($\beta$) & \textbf{82.5}(0.87) & \textbf{54.1}(0.95)  & \textbf{90.4}(0.95)  & \textbf{94.2}(0.98) & \textbf{94.8}(0.95) & \textbf{84.7}(0.91) & \textbf{89.7}(0.98)\\

\hline
\hline
MVCNN & - & 49.6 & \underline{89.4} & 93.9 & - & - & -\\ 
MGNC-CNN & - & 48.7 & 88.3 & \underline{94.1} & 95.5 & - & -\\ 
CNN-Rule & 81.7 & - & 89.3 & - & - & 85.3 & -\\ 
Semantic-CNN & 82.1 & 50.8 & 89.0 & 93.7 & 94.4 & \underline{86.0} & \underline{89.3}\\ 
combine-skip & 76.5 & - & - & 93.6 & 92.2 & 80.1 & 87.1\\ 
DSCNN & \underline{82.2} & 50.6 & 88.7 & 93.9 & \underline{95.6} & - & -\\ 
Paragraph Vector & 74.8 & 48.7 & 87.8 & 90.5 & 91.8 & 78.1 & 74.2\\ 
NBSVM & 79.4 & - & - & 93.2 & - & 81.8 & 86.3\\ 
Tree LSTM & - & \underline{51.0} & 88.0 & - & - & - & -\\ 
\hline
\end{tabular}

\caption{Effectiveness of GI-Dropout.
Dropout-same means dropping units with the same probability.
Results also include: MVCNN \cite{yin2015multichannel}, MGNC-CNN \cite{zhang2016mgnc}, CNN-Rule \cite{hu2016harnessing}, Semantic-CNN \cite{li2017initializing}, combine-skip \cite{kiros2015skip}, combine-skip \cite{kiros2015skip}, DSCNN \cite{zhang2016dependency}, Paragraph Vector \cite{le2014distributed}, NBSVM \cite{wang2012baselines} and Tree LSTM \cite{tai2015improved}.
}

\label{table: result}
\qquad
\end{small}
\end{table*}

\subsection{Effectiveness of GI-Dropout}
Results on 7 datasets are listed in Table \ref{table: result}.
Experiments show that the models with GI-Dropout outperform both CNN and self-attentive RNN baselines by a significant margin.

To test whether global information makes key contribution, we conduct another experiment in which all words are dropped according to the same probability at the GI-Dropout layer. 
Grid search method is used to find the best result which is listed in ``Dropout-same-prob'' row.

The one-layer CNN provides a very strong baseline.
The first line is the result of CNN-non-static model in \cite{kim2014convolutional}.
We reproduce the experiment results in ``CNN-baseline'' row.

Table \ref{table: result} shows that by simply dropping all the words according to the same probability, the model gains slight improvements against CNN baseline on all the datasets except in MPQA. Similarly, it achieves improvements compared with RNN baseline on most datasets.

By integrating our GI-Dropout mechanism, the model further improves the performance significantly on both CNN and RNN models. Compared with Dropout-same, there is a clear advantage that results on all of the datasets have been improved.

With the comparison between GI-Dropout and Dropout-same, we are convinced that GI-Dropout benefits from global information which provides explicit semantic information to guide the training process.

Even when compared with other models with complex architectures, GI-Dropout models achieve the best accuracy on most datasets, especially in SST-1 and SST-2.

\begin{table}[!t]
\renewcommand{\arraystretch}{1.3}
\centering
\begin{tabular}{c | c c}
\hline
\bfseries $\beta$ & CNN &  RNN\\
\hline
0.98 ($10^{-0.01}$) & 48.8 & 51.9 \\
0.95 ($10^{-0.02}$) & \textbf{49.0} & \textbf{54.1} \\
0.91 ($10^{-0.04}$) & 48.0 & 51.8 \\
0.87 ($10^{-0.06}$) & 48.1 & 52.4 \\
0.83 ($10^{-0.08}$) & 47.4 & 51.4 \\

\hline
\hline
\end{tabular}
\caption{$\beta$ and accuracy in SST-1. }
\label{table: SST-1}
\end{table}

\subsection{Further Analysis of Our Method}
With GI-Dropout, we drop words according to their importance scores. 
The higher score of a word, the greater chance it is to be ignored.
We further analyze why GI-Dropout works so well, and the relationship between $\beta$ and accuracy.

\textbf{GI-dropout helps models to learn inapparent features.}
\begin{table}[!t]
\renewcommand{\arraystretch}{1.3}
\centering
\begin{tabular}{c | c c}
\hline
\bfseries Top-k &  CNN baseline &  GI-Dropout in CNN\\
\hline
0 & 87.5 & 88.1  \\
50 & 87.1 & 87.9  \\
100 & 86.7 & 87.9  \\
200 & 86.1 & 87.5  \\
500 & 84.7 & 86.6  \\
1000 & 81.7 & 84.0 \\

\hline
\hline
\end{tabular}
\caption{Accuracy decline when removing top-k apparent words in SST-2. }
\label{table: decline-SST-2}
\end{table}
To test whether the method indeed helps models to learn the inapparent features, we conduct experiments where the top-k apparent words (with highest important scores) were removed from test cases in “SST-2”. 
Results are shown in Table \ref{table: decline-SST-2}.
We can observe that the CNN baseline model is more sensitive to the apparent features and GI-dropout can still have relatively good results even when we remove top 1000 apparent words. 
Thus, the model is supposed to pay more attention to the inapparent features with the help of GI-Dropout.

\textbf{GI-dropout helps models to reduce the overfitting for the apparent features.}
The frequent words can easily induce the model to focus on limited features and activate a part of units with large score. This can be seen by analyzing the cases which the proposed model makes a correct prediction and the baseline makes a incorrect prediction:

\textbf{(1)}  \textit{provide -lrb- s -rrb- nail-biting suspense and credible characters \underline{without} relying on technology-of-the-moment technique or \textbf{pretentious}\footnote{Words in \textbf{bold} denote the apparent features with high importance scores, e.g. ``pretentious" appears 159 times in \textbf{positive} texts and 5 in \textbf{negative} texts. Words with \underline{underline} represent unobvious features that also contribute to the class prediction.} dialogue.}

\textbf{(2)}  \textit{the screenplay \underline{sabotages} the movie's \textbf{strengths} at almost every juncture.} 

\textbf{(3)}  \textit{this is \underline{cool}, \underline{slick} stuff, ready to \underline{quench} the thirst of an audience that \textbf{misses} the summer blockbusters.} 

The baseline model is prone to focus only on the prominent features, e.g. the \textbf{``pretentious"} (negative) in case (1), \textbf{``strengths"} (positive) in case(2) and \textbf{``miss"} (negative) in case (3), and then make wrong predictions. Even though there are some important words indicating the opposite polarity, e.g. \textbf{``without"} in case (1), \textbf{``sabotages"} in case (2) , \textbf{``cool"}, \textbf{``slick"} and \textbf{``quench"} in case (3), the model can not make use of these features efficiently. 

By integrating our GI-dropout method, the model can learn not only the obvious features, e.g. \textbf{``strengths"}, but also the less obvious features e.g. \textbf{``sabotages"}. Thus, it makes correct predictions in all the above cases.

\textbf{The relationship between $\beta$ and accuracy.}
Another thing should be noticed is the value of $\beta$ in Equation \ref{new_nb}.
As shown in Figure \ref{fig:zipf}, the probability of a word and its rank follow Zipf's Law when $\beta$ is 0.95 in SST-1. 
Actually, for each dataset, there is an appropriate $\beta$ value for Equation \ref{new_nb} that can approximate the dropout probability and its rank with a Zipfian distribution. We assume that the $\beta$ setting in accord with Zipf's Law could have an important positive effect on the model performance. To examine this hypothesis, we further test the influences of different $\beta$ values on the CNN and RNN model. As expected, Table \ref{table: SST-1} shows that the models achieve the best results for both CNN and RNN in SST-1 with $\beta$ setting to 0.95.

\section{\large{Conclusion}}
This paper proposes GI-Dropout, a novel dropout method which utilizes global information and guides neural networks to extract not only obvious features but also unobvious features.

This idea is inspired by dropout in which units are dropped randomly in training according to the same probability. 
Unlike traditional dropout method, we aim to use global information to guide our dropout based on the importance of the words.

By integrating this mechanism, we encode global information explicitly into model via a novel Na\"ive Bayes Weighting method. We discover that model can be sensitive to some inapparent patterns, which is of great help to the classification. Experimental results demonstrate the effectiveness of GI-Dropout on multiple text classification tasks. In addition, our method requires few external resources and relatively small calculation. It is simple but effective and could be easily applied to other NLP tasks.

\section*{Acknowledgments}
This work was supported by National Natural Science Foundation of China (No. 61702047), Beijing Natural Science Foundation (No. 4174098), the Fundamental Research Funds for the Central Universities (No. 2017RC02), National Social Science Fund of China (No. 18CYY029) and China Postdoctoral Science Foundation funded project (No. 2018M630095).

\bibliography{conll2018}
\bibliographystyle{acl_natbib_nourl}

\end{document}